\documentclass[10pt,twocolumn,letterpaper]{article}

\usepackage{iccv}
\usepackage{times}
\usepackage{epsfig}
\usepackage{graphicx}
\usepackage{amsmath}
\usepackage{amssymb}
\usepackage{adjustbox}
\usepackage{array}
\usepackage{multirow}
\usepackage{subfigure}

\newcolumntype{R}[2]{%
    >{\adjustbox{angle=#1,lap=\width-(#2)}\bgroup}%
    l%
    <{\egroup}%
}
\newcommand*\rot{\multicolumn{1}{R{45}{1em}}}
\newcommand{\minisection}[1]{\vspace{0.04in} \noindent {\bf #1}\ \ }

\usepackage[pagebackref=true,breaklinks=true,colorlinks,bookmarks=false]{hyperref}

\iccvfinalcopy 

\def\iccvPaperID{1797} 

\ificcvfinal\fi

\begin{document} 

\title{Domain-adaptive deep network compression}
\author{Marc Masana, Joost van de Weijer, Luis Herranz \\
CVC, Universitat Aut\`onoma de Barcelona\\
Barcelona, Spain\\
{\tt\small \{mmasana,joost,lherranz\}@cvc.uab.es}
\and
Andrew D. Bagdanov\\
MICC, University of Florence \\
Florence, Italy\\
{\tt\small andrew.bagdanov@unifi.it}
\and
Jose M \'Alvarez\\
Toyota Research Institute\\
{\tt\small jose.alvarez@tri.global}
}

\maketitle

\begin{abstract} 
Deep Neural Networks trained on large datasets can be easily transferred to new domains with far fewer labeled examples by a process called fine-tuning. This has the advantage that representations learned in the large source domain can be exploited on smaller target domains. However, networks designed to be optimal for the source task are often prohibitively large for the target task. In this work we address the compression of networks after domain transfer. 

We focus on compression algorithms based on low-rank matrix decomposition. Existing methods base compression solely on learned network weights and ignore the statistics of network activations. We show that domain transfer leads to large shifts in network activations and that it is desirable to take this into account when compressing. We demonstrate that considering activation statistics when compressing weights leads to a rank-constrained regression problem with a closed-form solution. Because our method takes into account the target domain, it can more optimally remove the redundancy in the weights. Experiments show that our Domain Adaptive Low Rank (DALR) method significantly outperforms existing low-rank compression techniques. With our approach, the fc6 layer of VGG19 can be compressed more than 4x more than using truncated SVD alone -- with only a minor or no loss in accuracy. When applied to domain-transferred networks it allows for compression down to only 5-20\% of the original number of parameters with only a minor drop in performance.
\end{abstract}

\begin{figure}[ht]
\centering
\includegraphics[width=8.0cm]{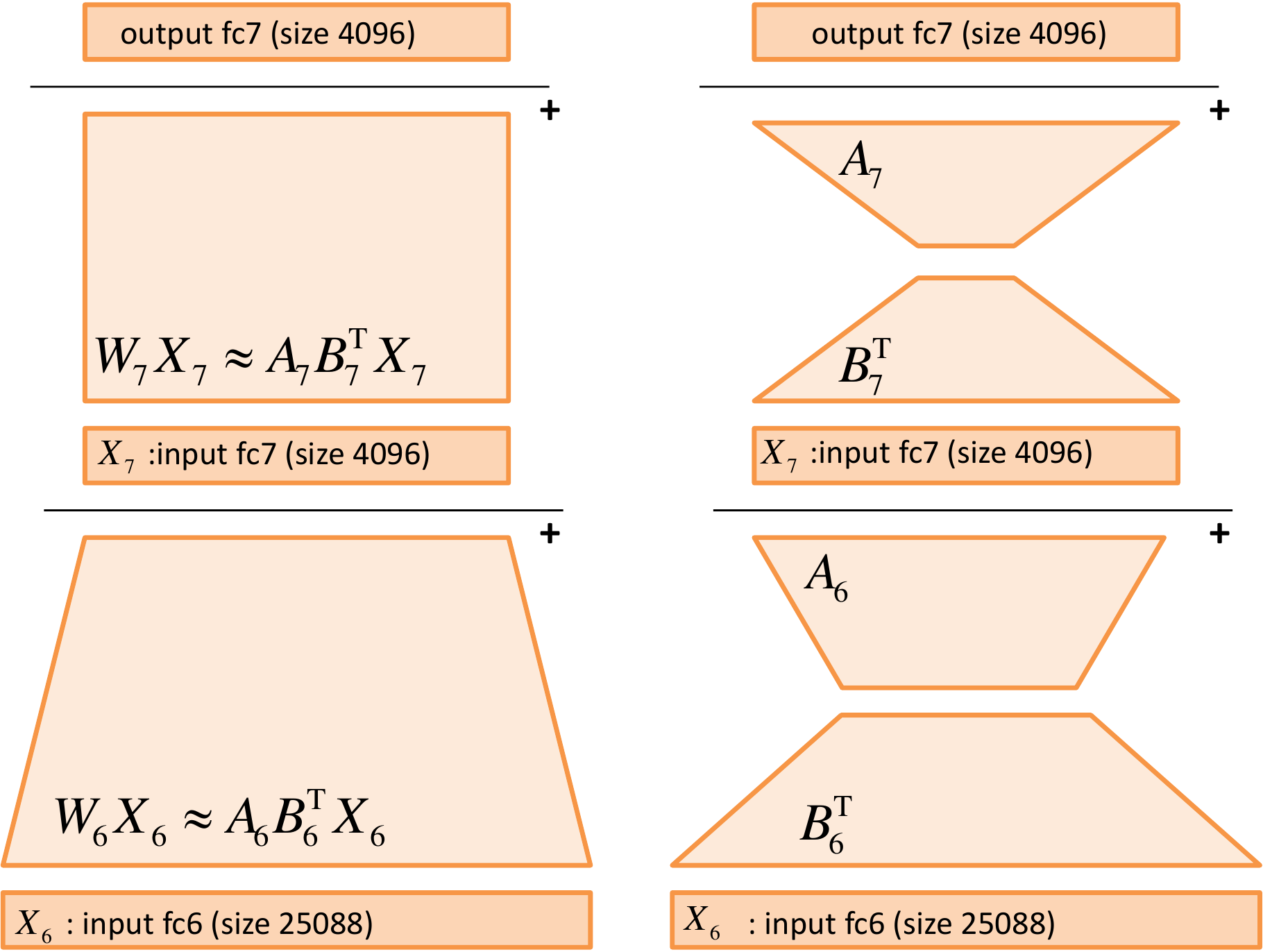}
\caption{\textit{}Example of compressing last two layers of the VGG-16 network (fc6, fc7). The original weight matrix is approximated by two matrices. The main novelty in this paper is that we consider the input $X$ of each layer when compressing the corresponding weight matrix $W$. This is especially relevant when doing domain transfer from pre-trained networks where activation statistics can be significantly skewed in the target domain.}
\label{fig_decom3}\label{fig_decom2}
\end{figure}

\section{Introduction}\label{sec:introduction}
One of the important factors in the success of deep learning for computer vision is the ease with which features, pre-trained on large datasets such as Imagenet~\cite{deng2009imagenet,russakovsky2015imagenet} and
Places2~\cite{zhou2016places}, can be transferred to other computer vision domains. These new domains often have far fewer labeled samples available but which, due to the high correlation which exists between visual data in general, can exploit an already learned representation trained on large datasets. The most popular method to transfer the representations is by means of fine-tuning, where the network is initialized with the pre-trained network weights, after which they are further adjusted on the smaller target domain~\cite{oquab2014learning}. These fine-tuned networks, which have the same size as the originally trained network, can then be applied to the task of the target domain. However, one must question whether a target domain task requires such a large network and whether the resulting network is not highly redundant.

A drawback of Convolutional Neural Networks (CNNs) is that they generally require large amounts of memory and computational power (often provided by GPU). As a result they are less suitable for small devices, like cell phones, where requirements for energy efficiency limit CPU, GPU, and memory usage. This observation has motivated much research into network compression. Approaches include methods based on weight quantization~\cite{han2015deep,rastegari2016xnor}, weight removal from fully-connected~\cite{zhou2016less} or convolutional layers~\cite{AlvarezNIPS}, compact representations of convolutional layers through tensor decompositions~\cite{denton2014exploiting,jaderberg2014speeding,alvarez2016decomposeme}, as well as training of thinner networks from predictions of a larger {\it teacher} network~\cite{Hinton06,romero2014fitnets}.

One efficient method for the compression of fully connected layers is based on applying singular value decomposition (SVD) to the weight matrix~\cite{Jinyu:2013,denton2014exploiting,zhou2016less}. Compression is achieved by removing columns and rows related to the least significant singular values. Then, the original layer is replaced by two layers which have fewer parameters than the original layer. The method has been successfully applied to increase efficiency in detection networks like Fast R-CNN~\cite{girshick2015fast}. In these networks the truncated SVD approach is applied to the fc6 and fc7 layers, and the authors showed that with only a small drop in performance these layers can be compressed to 25\% of their original sizes. In the original paper~\cite{Jinyu:2013, denton2014exploiting, zhou2016less} the compression is always applied on the source domain, and no analysis of its efficiency for domain transfer exists. 

In this work we investigate network compression in the context of domain transfer from a network pre-trained on a large dataset to a smaller dataset representing the target domain. To the best of our knowledge we are the first to consider network compression within the context of domain transfer, even though this is one of the most common settings for the application of deep networks. Our approach is based on weight matrix decomposition that takes into account the activation statistics of the original network on the target domain training set\footnote{With activation statistics we refer to their direct usage during compression, but we do not explicitly model the statistical distribution.}. We first adapt a pre-trained network with fine-tuning to a new
target domain and then proceed to compress this network. We argue that, because the statistics of the activations of network layers change from the source to the target domain, it is beneficial to take this shift into account. Most current compression methods do not consider activation statistics and base compression solely on the values in the weight matrices~\cite{lecun1989optimal, Jinyu:2013, denton2014exploiting, han2015deep, rastegari2016xnor}. We show how the activation statistics can be exploited and that the resulting minimization can be written as a rank-constrained regression problem for which there exists a closed-form solution. We call our approach Domain Adaptive Low Rank (DALR) compression, since it is a low-rank approximation technique that takes into account the shift in activation statistics that occurs when transferring to another domains. As an additional contribution, we show that the original SVD algorithm can be improved by compensating the bias for the activations.

The paper is organized as follows. In the next section we discuss work from the literature related to network compression. In section~\ref{sec:motivation} we discuss in detail the motivation behind our network compression approach, and in section~\ref{sec:compression} we show how network compression can be formulated as a rank constraint regression problem. In section~\ref{sec:experiments} we report on a range of compression experiments performed on standard benchmarks. Finally, we conclude with a discussion of our contribution in section~\ref{sec:conclusions}.

\section{Related Work}
\label{sec:related_work}
Network compression has received a great deal of attention recently. In this section we briefly review some of the works from the literature relevant to our approach.

\minisection{Network pruning.} A straight forward way to reduce the memory footprint of a neural network is by removing unimportant parameters. This process can be conducted while training~\cite{AlvarezNIPS, zhou2016less,BrainSurgeon,lecun1989optimal}, or by analyzing the influence of each parameter once the network has been trained~\cite{Liu:CVPR2015_SparseNets}. For instance, in \cite{zhou2016less}, the authors use tensor low rank constraints to (iteratively) reduce the number of parameters of the fully connected layer. 

\minisection{Computationally efficient layer representations.} Several approaches have addressed the problem of reducing computational resources by modifying the  internal representation of each layer taking into account the inherent redundancy of its parameters. Common approaches exploit linear structures within convolutional layers and approach each convolutional kernel using low-rank kernels~\cite{denton2014exploiting,jaderberg2014speeding,alvarez2016decomposeme}. The main idea relies on the fact that performing a convolution with the original kernels is equivalent to convolving with a set of base filters followed by a linear combination of their output. In~\cite{Corso:2016}, the authors propose two network layers that are based on dictionary learning to perform sparse representation learning, directly within the network. In general, these approaches show significant reduction in the computational needs with a minimum drop of performance.

\minisection{Parameter quantization.} Previous works mentioned above on efficient representations focus on modifying the representation of a complete layer. Parameter quantization is slightly different as it aims at finding efficient representations of each parameter individually (ie, representing each parameter with fewer bits). A common practice to minimize the memory footprint of the network and reduce the computational cost during inference consists of training using $32$ bit to represent the parameters while performing inference more efficiently using 16-bits without significantly affecting the performance. More aggressive quantization processes have been also analyzed in~\cite{GongCompression} where the authors propose an approach to directly thresholding values at 0 resulting in a decrease of the top-1 performance on ImageNet by less tan 10\%. More recently, several works have adopted quantization schemes into the training process~\cite{rastegari2016xnor}. For instance, in~\cite{rastegari2016xnor}, the authors propose an approach to train a network directly using binary weights resulting in very efficient networks with a very limited accuracy.

In~\cite{han2015deep} the authors propose an approach to combine pruning, quantization and coding to reduce the computational complexity of the network. The authors first analyze the relevance of each parameter pruning the irrelevant ones and then, after fine-tuning the pruned network, the remaining parameters are quantized. Results show a significant reduction in the number of parameters (up to 35x) without affecting the performance of the network. 

\minisection{Network distillation.} These approaches aim at mimicking a complicated model using a simpler one. The key idea consists of training an ensemble of large networks and then use their combined output to train a simpler model~\cite{BucilaCN06}. Several approaches have built on this idea to train the network based on the soft output of the ensemble~\cite{distNets}, or to train the network mimicking the behavior not only of the last layer but also of intermediate ones~\cite{romero2014fitnets}.

All these methods for pruning, quantization, or compression in general, have shown results for reducing the footprint of networks and for reducing its computational complexity. However, they are usually applied to the same target domain as the one used for training the original network. In contrast, in this paper we investigate network compression in the context of domain transfer. That is, compressing a network that has been trained on a generic large dataset in order to reduce its computational complexity when used in a different target domain using a smaller dataset. In this context, the most related work is the approach presented in~\cite{gui:2015} exploring non-negative matrix factorization to identify an interesting set of variables for domain transfer. However, the work in~\cite{gui:2015} does not consider network compression and focuses on unsupervised tasks. 

\section{Motivation}
\label{sec:motivation}
Training networks for a specific task starting from a network pre-trained on a large, generic dataset has become very common practice, and to the best of our knowledge we are the first to consider compression of these types  of domain-adapted networks. We investigate the compression of fully connected layers by means of matrix decomposition. The basic principle is to decompose a weight matrix into two matrices which contain fewer parameters than the original weight matrix. This decomposition can then be applied to the network by replacing the original layer with two new layers (see Fig.~\ref{fig_decom2}). An existing approach is based on truncated
Singular Value Decomposition~\cite{Jinyu:2013,denton2014exploiting}. The decomposition of that method is determined solely by the weight matrix and ignores the statistics of layer activations in the new domain.

\begin{figure}[t]
  \centering
  \begin{tabular}{c}
      \includegraphics[width=0.95\columnwidth]{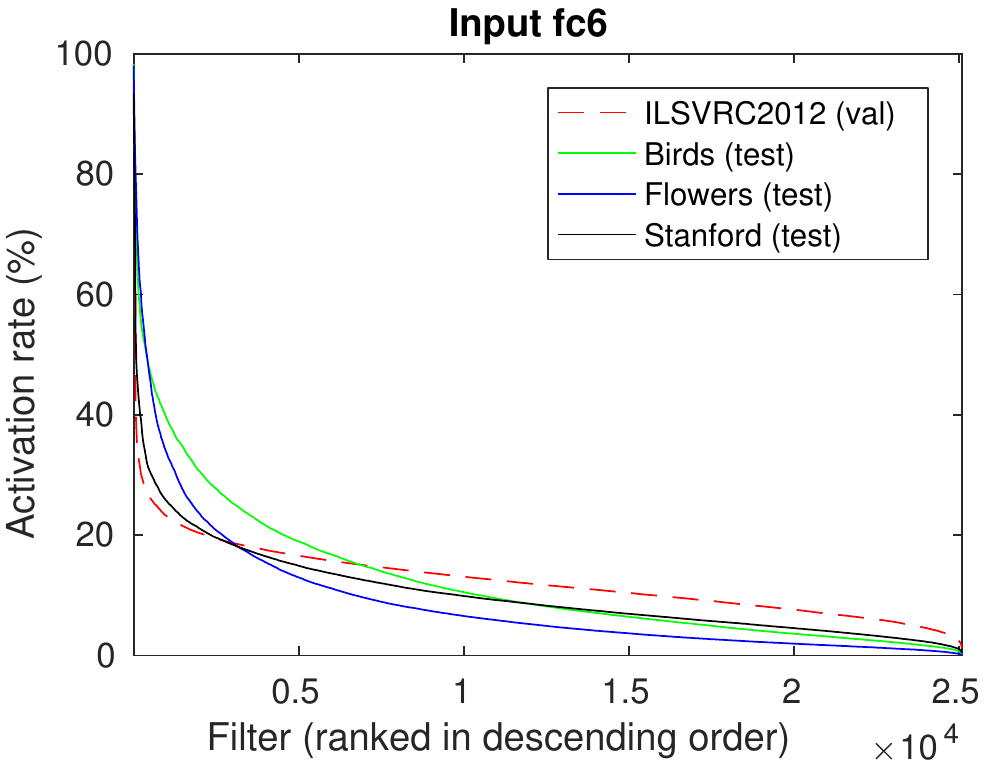} \\
 \includegraphics[width=0.95\columnwidth]{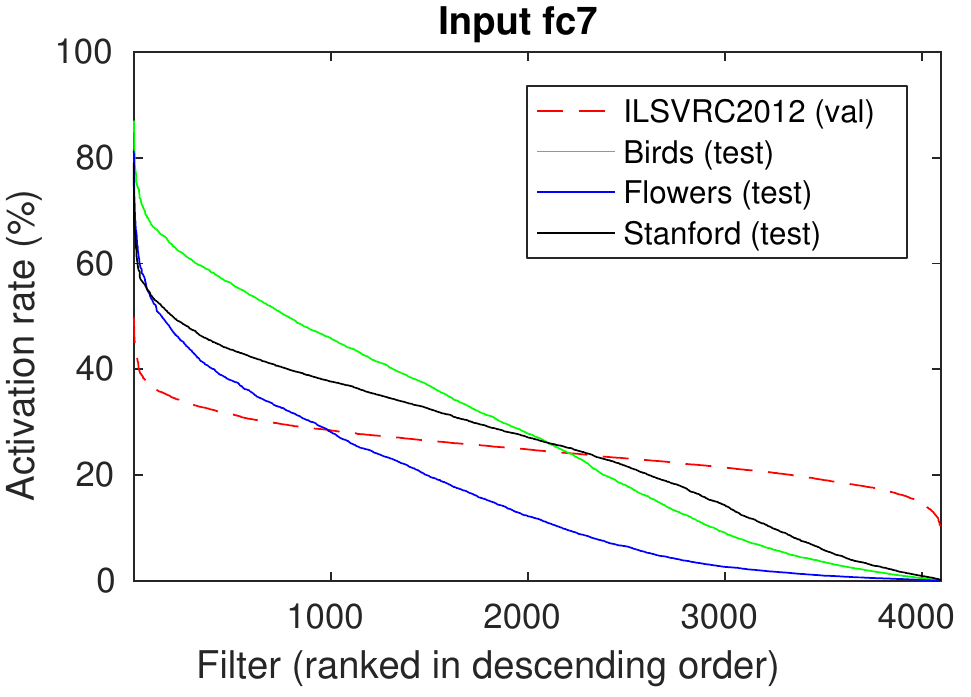}
  \end{tabular}
  \caption{Activation rates (ranked in decreasing order) of the input to (top) fc6, and (bottom) fc7 for the Birds, Flowers and Stanford datasets in the VGG19 trained on ILSVRC2012. The dimensions of the inputs are $7\times7\times512=25088$ (output of pool5) and 4096 (output of fc6), respectively. Note the more uniform distribution for ILSVRC2012 (no domain change). Best viewed in color.
    }
    \label{fig:activation_rates}
\end{figure}

To gain some insight into this shifting of the activation distributions in deep networks when changing domain, we take a closer look at the inputs to the two fully connected layers fc6 and fc7 (which are the output of pool5 and fc6, respectively), of the VGG19 network~\cite{simonyan2014very}. We analyze the {\it activation rate} of neurons which is the fraction of images in which a neuron has non-zero response. A value of 0.3 means that the neuron is activated in 30\% of the images in the data set. In Fig.~\ref{fig:activation_rates} we show the activation rates of the VGG19 network on ImageNet, on the CUB-200-2011 Birds dataset~\cite{WahCUB_200_2011}, on the Oxford 102 Flowers dataset~~\cite{Nilsback08} and on the Stanford 40 actions dataset~\cite{yao2011human}.

The first thing to observe is that the activation rate is fairly constant across all the input dimensions (i.e. activation rate of neurons in the previous layer) when computed on the ImageNet dataset (i.e. source domain). Apparently the network has optimized its efficiency by learning representations which have a similar frequency of occurrence in the ImageNet dataset. However, if we look at the activation rates in the three target domains we see that the distributions are significantly skewed: a fraction of neurons is much more frequent than the rest, and the activation rates are lower than in the source domain. This is especially clear for the input to fc7 where activation rates vary significantly. If we consider the  percentage of input dimensions which accumulates 50\% of the activations (which is the point where the area under the curve to the left is equal to the area under the curve to the right), we see a clear shift from ImageNet with 41.38\% to 19.51\% in Flowers, 24.93\% in Birds and 29.44\% in Stanford (and from 32.29\% to 14.61\%, 19.13\% and 25\% for fc6, respectively). This clearly shows that there exists a significant change in the relative importance of neurons from previous layers, optimized for a source domain, when applied on new domains. Given this significant shift, we believe that it is important to take these activation statistics into account when compressing network layers after domain transfer. Keeping lower weights connected to high activation neurons can lead to more efficient compression rather than only focusing on the value of the weights themselves as is done by current compression methods.

\section{Compression by matrix decomposition}
\label{sec:compression}
We start by introducing some notations. Consider a single fully connected layer, with input and bias
$x,b \in \mathbb{R}^n$, the output $y \in \mathbb{R}^m$ and the layer weights $W \in \mathbb{R}^{m\times n}$, related according to:
\begin{equation}
y = Wx + b,
\end{equation}
or when considering a set of $p$ inputs to the layer:
\begin{equation}
Y=WX + b1^T_p,
\end{equation}
where $1_p$ is a vector with ones of size $p\times 1$, $Y \in \mathbb{R}^{m\times p}$ is the set of outputs, and $X \in \mathbb{R}^{n\times p}$ is the set of $p$ inputs to the layer.

Several compression works have focused on compressing $W$ into $\hat W$ so that $||W - \hat W||_F$ is minimized ~\cite{lecun1989optimal,Jinyu:2013,denton2014exploiting}. The novelty of our work is that we also consider the inputs $X$. As a consequence we will focus on  compressing $W$ into $\hat W$ in such a way that $||Y - \hat Y||_F$ is minimal.

\subsection{Truncated SVD and Bias Compensation (BC)}
One approach to compression is to apply SVD such that $W = USV^T$ where $U \in \mathbb{R}^{m \times m}$, $S \in \mathbb{R}^{m \times n}$, $V \in \mathbb{R}^{n \times n}$~\cite{lecun1989optimal,Jinyu:2013}. The layer weights $W$ can be approximated by keeping the $k$ most significant singular vectors, $\hat W = \hat U\hat S\hat V^T$ where $\hat U \in \mathbb{R}^{m \times k}$, $\hat S \in \mathbb{R}^{k \times k}$, $\hat V \in \mathbb{R}^{n \times k}$.  Compression is obtained by replacing the original layer by two new ones: the first with weights $\hat S\hat V^T$ and the second with weights $\hat U$. Note it is crucial that the two new layers contain fewer parameters than the original network (i.e. that $nm > (n + m)k$). 

In this truncated SVD approach the bias term $b$ of the original network is added to the second layer. We propose an alternative bias term which takes into account the inputs $X$. We define $
W = \hat W +  \mathord{\buildrel{\lower3pt\hbox{$\scriptscriptstyle\smile$}} \over W} $, where $\mathord{\buildrel{\lower3pt\hbox{$\scriptscriptstyle\smile$}} \over W}$ is the residual which is lost due to the approximation. We want to find the new bias that minimizes $||Y - \hat Y||_F$ given  inputs $X$. Accordingly: 
\begin{eqnarray}
\nonumber
|| Y - \hat Y ||_F &=& || WX + b1_p^T  - ( {\hat WX + \hat b1_p^T } ) ||_F \\
&=& || \hat b1_p^T  - ( {b1_p^T  + \mathord{\buildrel{\lower3pt\hbox{$\scriptscriptstyle\smile$}} 
\over W} X} ) ||_F.
\end{eqnarray}
The bias which minimizes this is then:
\begin{equation}
\hat b = b + \mathord{\buildrel{\lower3pt\hbox{$\scriptscriptstyle\smile$}} 
\over W} X1_p  = b + \mathord{\buildrel{\lower3pt\hbox{$\scriptscriptstyle\smile$}} 
\over W} \bar x \label{Eq_bias}
\end{equation}
where $\bar x= X1_p$ is the mean input response. Note that if $X$ were zero centered, then  $\bar x$ would be zero and the optimal bias $\hat b = b$. However, since  $X$ is typically taken right after a ReLU layer, this is generally not the case and SVD can introduce a systematic bias in the output which in our case can be compensated for using Eq.~\ref{Eq_bias}.

\subsection{Domain Adaptive Low Rank Matrix Decomposition (DALR)}
In the previous subsection we considered the inputs to improve the reconstruction error by compensating for the shift in the bias. Here, we also take into account the inputs for the computation of the matrix decomposition. In contrast, the SVD decomposition does not take them into account. Especially in the case of domain transfer, where the statistics of the activations can significantly differ between source and target domain, decomposition of the weight matrix should consider this additional information. A network trained on ImageNet can be highly redundant when applied to for example a {\it flower} dataset; in this case most features important for man-made objects will be redundant.

The incorporation of input $X$ is done by minimizing $|| Y - \hat Y||_{F}$. We want to decompose the layer with weights $W$ into two layers according to:
\begin{equation}
W \approx \hat W = AB^T, 
\label{eq_lowrank}
\end{equation}
where $\hat W \in {R}^{m\times n}$,  $A \in {R}^{m \times k}$ and $B \in {R}^{n \times k}$ again chosen in such a way that $m \times n > (m+n) \times k $.

We want the decomposition which minimizes: 
\begin{equation}
\min_{A,B} || Y - \hat Y ||_F = \min_{A,B} || WX - AB^T X ||_F, \label{eq_layerdec}
\end{equation}
where we have set $\hat b=b$ and subsequently removed it from the equation. Eq.~\ref{eq_layerdec} is a rank constrained regression problem which can be written as:
\begin{equation}
\begin{array}{l}
 \mathop {\arg \min }\limits_C || Z - CX ||_F^2 {\rm{ + }}\lambda ||C||_F^2 \\ 
 \mathrm{s.t. \ rank}(C) \le k, \\ 
 \end{array}
 \label{eq:rank-regression}
\end{equation}
where $C=AB^T$ and $Z=WX$, and we have added a ridge penalty which ensures that $C$ is well-behaved even when $X$ is highly co-linear.

We can rewrite Eq~\ref{eq:rank-regression} as: 
\begin{equation}
\begin{array}{l}
  \mathop{\arg \min}\limits_C || Z^* - CX^* ||_F^2 \\ 
 \mathrm{s.t. \ rank}(C) \le k, \\ 
 \end{array}\label{eq_rrr}
\end{equation}
where we use 
\begin{eqnarray}
X^*_{n\times(p+n)} &=& \left({\begin{array}{*{20}c}
   X & {\sqrt \lambda  I}  \\
\end{array}} \right), \mathrm{ and}\\
Z^*_{m\times(p+n)} &=& \left( {\begin{array}{*{20}c}
   Z & 0  \\
\end{array}} \right).
\end{eqnarray}

In Ashin~\cite{mukherjee2013topics} the authors show that there is a closed form solution for such minimization problems based on the SVD of $Z$. Applying SVD we obtain $Z=USV^T$. Then the matrices A and B in Eq.~\ref{eq_lowrank} which minimize Eq.~\ref{eq_rrr} are:
\begin{equation}
\begin{array}{l}
 A = \hat U \\ 
 B = \hat U^T ZX^T \left( {XX^T  + \lambda I} \right)^{ - 1}  \\ 
 \end{array}
\end{equation}
where $\hat U \in {R}^{m \times k}$ consists of the first $k$ columns of $U$.

Network compression is obtained by replacing the layer weights $W$ by two layers with weights $B$ and $A$, just as in the truncated SVD approach. The first layer has no biases and the original biases $b$ are added to the second layer. Again we could apply Eq.~\ref{Eq_bias} to compensate the bias for the difference between $W$ and $\hat W$. However, this was not found to further improve the results. 

\subsection{Reconstruction error analysis}
\label{sec:reconstruction}
We discussed three different approaches
to compressing a weight matrix $W$. They lead to the following approximate
outputs $\hat Y$:
\begin{eqnarray}
  \nonumber
  \mathrm{SVD}:      \hat Y &=& \hat U\hat S\hat V^{\rm{T}} X + b \\ 
  \nonumber
  \mathrm{SVD + BC}: \hat Y &=& \hat U\hat S\hat V^{\rm{T}} X + \hat b \\ 
  \mathrm{DALR}:     \hat Y &=& AB^T X + b
\end{eqnarray}
To analyze the ability of each method to approximate the original output
$Y$ we perform an analysis of the reconstruction error given by:
\begin{equation}
\varepsilon  = || Y - \hat Y ||_F.
\end{equation}
We compare the reconstruction errors on the CUB-200-2011 Birds and
the Oxford-102 Flowers datasets.  The reconstruction error is evaluated on
the test set, whereas the inputs and outputs of the layers are extracted
from the training set for computing the matrix approximations
$\hat Y$. We provide results for fc6 and fc7, the two fully connected
layers of the VGG19 network.

\begin{figure}[t]
    \centering
    \subfigure{\includegraphics[width=0.97\columnwidth]{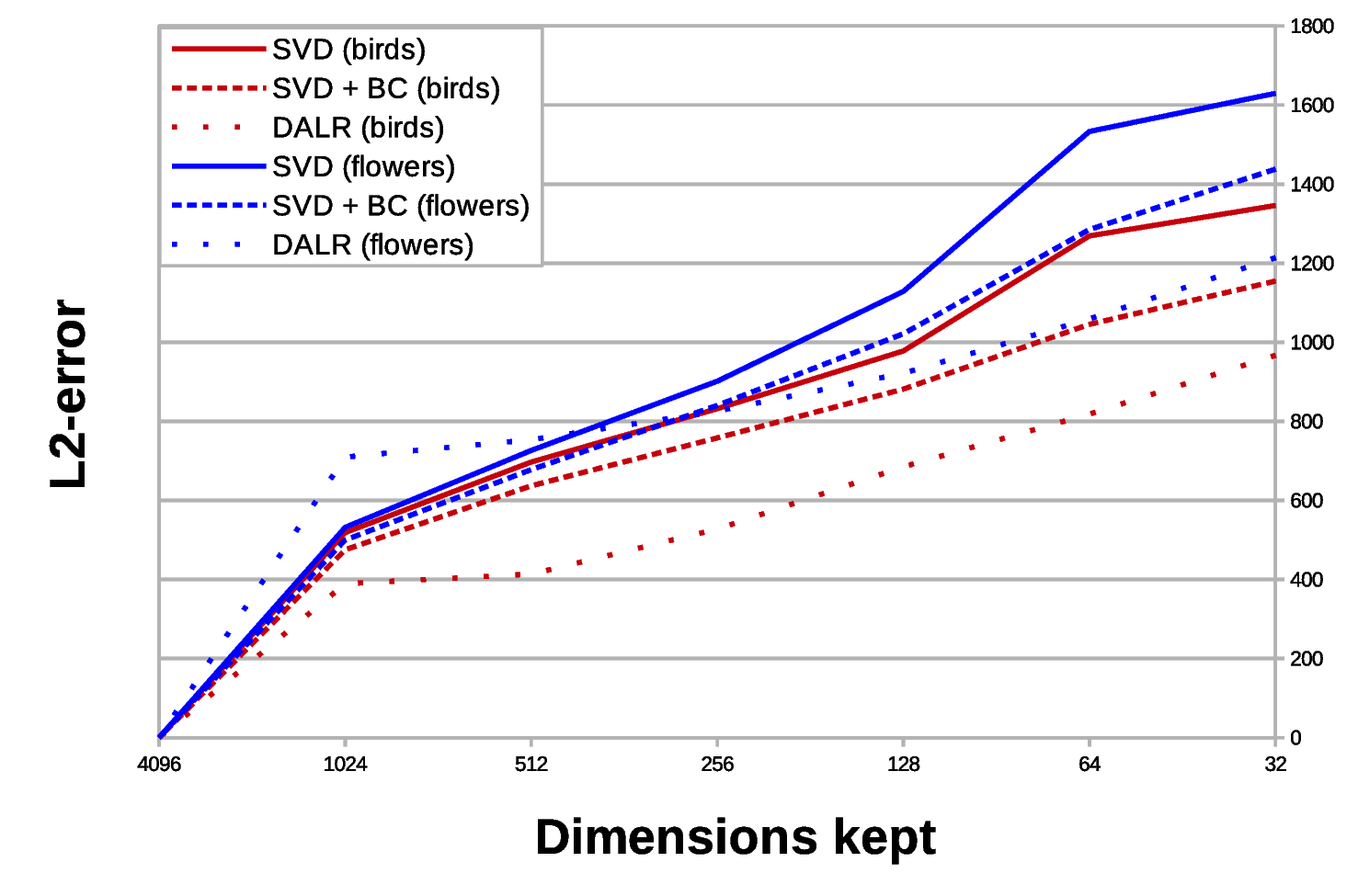}}
    \subfigure{\includegraphics[width=0.97\columnwidth]{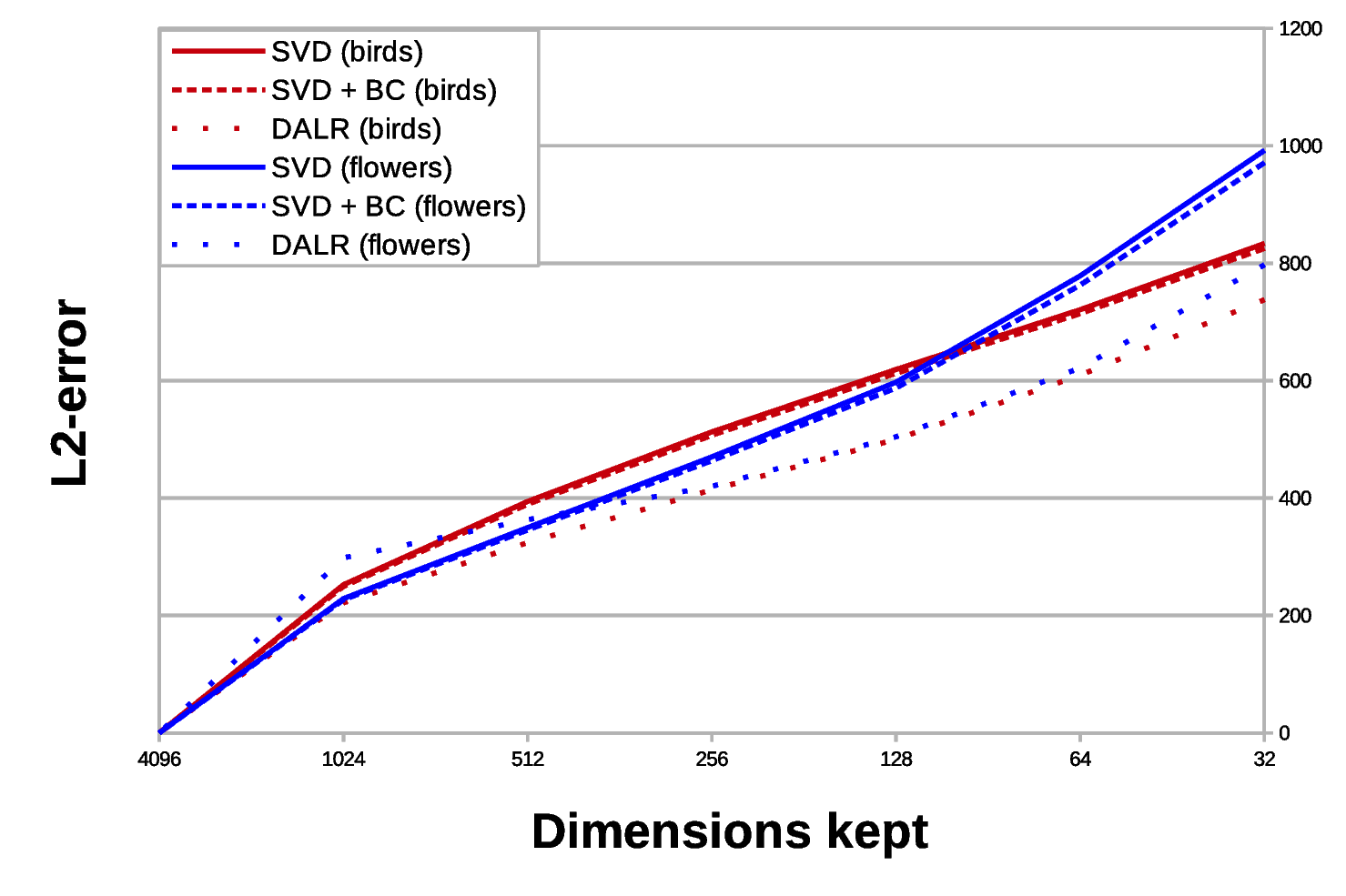}}
    \caption{Reconstruction error as a function of dimensions kept $k$ for (top) fc6 and (bottom) fc7 layers on CUB-200-2011 Birds and Oxford-102 Flowers depending on the degree of compression.}
    \label{fig:L2_error_reconstr}
\end{figure}

In Figure~\ref{fig:L2_error_reconstr} we show the results of
this analysis. We see that bias compensation provides a drop in error with respect to SVD for layer fc6, however the gain is insignificant for layer fc7. Our DALR method obtains lower errors for both layers on both datasets for most of the compression settings. This shows the importance of taking activation statistics into account during compression.

\section{Experimental results}
\label{sec:experiments}
Here we report on a range of experiments to quantify the effectiveness of our network compression strategy. Code is made available at \href{https://github.com/mmasana/DALR}{https://github.com/mmasana/DALR}.

\subsection{Datasets}
  We evaluate our DALR approach to network compression on a number of standard datasets for image recognition and object detection.
  
\minisection{CUB-200-2011 Birds:} consists of 11,788 images (5,994 train) of 200 bird species~\cite{WahCUB_200_2011}. Each image is annotated with bounding box, part location (head and body), and attribute labels. Part location and attributes are not used in the proposed experiments. However, bounding boxes are used when fine-tuning the model from VGG19 pre-trained on ImageNet in order to provide some data augmentation for the existing images.

\minisection{Oxford 102 Flowers:} consists of 8,189 images (2,040 train+val) of 102 species of flowers common in the United Kingdom~\cite{Nilsback08}. Classes are not equally represented across the dataset, with the number of samples ranging from 40 to 258 per class.

\minisection{Stanford 40 Actions:} consists of 9,532 images (4,000 for training) of 40 categories that depict different human actions~\cite{yao2011human}. Classes are equally represented on the training set and all samples contain at least one human performing the corresponding action.

\minisection{PASCAL 2007:} consists of approximately 10,000 images (5,011 train+val) containing one or more instances of 20 classes~\cite{everingham2010pascal}. The dataset contains 24,640 annotated objects, with that the training and validation sets having a mean of 2.52 objects per image, and the test set a mean of 2.43 objects per image. This dataset is used to evaluate our approach for object detection.

\minisection{ImageNet:} consists of more than 14 million images of thousands of object classes. The dataset is organized using the nouns from the WordNet hierarchy, with each class having an average of about 500 images. We use CNNs pre-trained on the 1,000-class ILSVRC subset.

\subsection{Compression on source domain}
Before performing experiments on domain adaptation, we applied Bias Compensation and DALR compression on the source domain and compare it to the truncated SVD method as baseline. We used the original VGG19 network trained on the 1,000 ImageNet classes. Since ImageNet is a very large dataset, we randomly selected 50 images from each class to build our training set for DALR and truncated SVD compression. Then, we extracted the activations of this training set for the fc6 and fc7 layers to compress the network using DALR at different rates. The compressed networks were evaluated on the ImageNet validation set.

Results are provided in Tables~\ref{imagenetfc6} and~\ref{imagenetfc7}. Results on fc6 show a slight performance increase for the most restricting compression settings ($k=32,64,128$). However the gain is relatively small. On fc7 the results are even closer and the best results are obtained with bias compensation. Even though the proposed compression methods outperform standard SVD the gain is small when done on the source domain. This is most probably due to the fact that the inputs have relatively uniform activation rates and considering them does not significantly change the matrix decomposition (see also the discussion in Section~\ref{sec:reconstruction}). In general these results suggest that compressing a model without changing the domain can be done effectively with decompositions of filter weights (e.g truncated SVD), and does not benefit significantly from the additional information coming from the inputs, in contrast to when there is a domain change involved (see the following sections).

\begin{table}
\resizebox{\linewidth}{!}{%
\begin{tabular}{lcccccc}
\hline
dim kept             & 32     & 64     & 128    & 256    & 512     & 1024    \\
params               & 0.91\% & 1.82\% & 3.64\% & 7.27\% & 14.54\% & 29.08\% \\ \hline
SVD                  & 79.44 & 50.82 & 36.44 & 34.80 & 34.40 & \textbf{34.18} \\
SVD + BC             & 73.41 & 46.54 & 36.21 & 34.82 & 34.33 & 34.22 \\
DALR                 & \textbf{66.43} & \textbf{44.50} & \textbf{36.06} & \textbf{34.63} & \textbf{34.28} & 34.20 \\ \hline
\end{tabular}%
}
\caption{Top-1 error rate results on ImageNet when compressing fc6 in the source domain. We report the dimensions $k$ kept in the layer and the percentage of parameters compressed. The uncompressed top-1 error rate is 34.24\%.}
\label{imagenetfc6}
\end{table}

\begin{table}
\resizebox{\linewidth}{!}{%
\begin{tabular}{lcccccc}
\hline
dim kept             & 32     & 64     & 128    & 256    & 512   & 1024  \\
params               & 1.56\% & 3.13\% & 6.25\% & 12.5\% & 25\%  & 50\%  \\ \hline
SVD                  & 57.07 & 40.68 & 35.50 & 34.63 & 34.40 & 34.35 \\
SVD + BC             & \textbf{55.57} & \textbf{39.75} & \textbf{35.14} & \textbf{34.51} & \textbf{34.35} & \textbf{34.30} \\
DALR                 & 56.32 & 40.25 & 35.26 & 34.54 & 34.40 & 34.33 \\ \hline
\end{tabular}%
}
\caption{ImageNet fc7 - uncompressed top-1 error rate: 34.24\%.}
\label{imagenetfc7}
\end{table}

\subsection{Image recognition}
Here we use the CUB-200 Birds, Oxford-102  Flowers and Stanford 40 Actions datasets to evaluate our compression strategies. We apply the Bias Compensation and the DALR compression techniques to fine-tuned VGG19 models. For all image recognition experiments we used the publicly available MatConvNet library~\cite{vedaldi15matconvnet}.

\minisection{Fine-tuning:} The VGG19~\cite{simonyan2014very} is trained on ImageNet. Since this network excelled on the ImageNet Large-Scale Visual Recognition Challenge in 2014 (ILSVRC-2014), it is a strong candidate as a pre-trained CNN source for the transfer learning. Very deep CNNs are commonly used as pre-trained CNNs in order to initialize the network parameters before fine-tuning. For each dataset, a fine-tuned version of VGG19 was trained using only the training set. Although initialized with the VGG19 weights, layers fc6 and fc7 are given a 0.1 multiplier to the network's learning rate. The number of outputs of the fc8 layer is changed to fit the number of classes in the dataset. All the convolutional layers are frozen and use the VGG19 weights.

\minisection{Evaluation metrics:} All results for image recognition are reported in terms of classification accuracy. The compression rate of fully connected layers is the percentage of the number of parameters of the compressed layer with respect to the original number of parameters.

\minisection{Baseline performance:} We first apply truncated SVD to the fc6 and fc7 weight matrices. In the original VGG19 and fine-tuned models, $W_{fc6}$ has $25088\times4096$ parameters and $W_{fc7}$ has $4096\times4096$ parameters. Applying truncated SVD results in a decomposition of each weight matrix into the two smaller matrices. If we keep the $k$ largest singular vectors, those two matrices will change to $(25088+4096)k$ and $(4096+4096)k$ parameters for fc6 and fc7 respectively. Since SVD does not take into account activations, and there is no compression method to our knowledge that uses activations in order to reduce the number of parameters in the weight matrices, we also show results for activation-based pruning. The pruning strategy consists of removing the rows or columns of the weight matrices which are less active for that specific dataset, following the work on~\cite{masana2016fly}.

\minisection{Results:}
Tables~\ref{birdsfc6} to~\ref{stanfordfc7} show the performance of compressing the fc6 and fc7 layers using SVD and pruning baselines, as well as the proposed Bias Compensation and DALR techniques. Results confirm the tendency observed in the analysis of the L2-error reconstruction curves in Figure~\ref{fig:L2_error_reconstr}. DALR compression has a better performance than the other methods at the same compression rates on both fc6 and fc7 for CUB-200 Birds, Oxford-102 Flowers and Stanford-40 Actions. In all our experiments DALR provides a slight boost in performance even when compressing to 25\% of the original parameters. Bias compensation slightly improves the original SVD method on both layers except on Flowers for fc7. Since the fc6 layer has more parameters, it is the layer that allows for more compression at a lower loss in performance. The advantages of DALR are especially clear for that layer, and for a typical setting where one would accept a loss of accuracy of around one percent, truncated SVD must retain between 4x and 8x the number of parameters compared to DALR to maintain the same level of performance. Finally, both pruning methods are consistently outperformed by compression methods, probably due to the effect pruning has on subsequent layers (fc7 and fc8).

\begin{table}
\resizebox{\linewidth}{!}{%
\begin{tabular}{lcccccc}
\hline
dim kept             & 32     & 64     & 128    & 256    & 512     & 1024  \\
params               & 0.91\% & 1.82\% & 3.64\% & 7.27\% & 14.54\% & 29.08\% \\ \hline
SVD                  & 16.83 & 36.47 & 51.74 & 54.47 & 54.85 & 55.25 \\
SVD + BC             & 27.91 & 46.00 & 53.50 & 54.83 & 55.21 & 55.44 \\ \hline
Pruning (mean)       &  4.30 &  8.06 & 12.57 & 25.82 & 37.25 & 50.41 \\ 
Pruning (max)        &  4.06 &  7.01 & 15.26 & 25.27 & 36.69 & 48.95 \\ \hline
DALR                 & \textbf{48.81}  & \textbf{54.51}  & \textbf{55.78}  & \textbf{55.85}  & \textbf{55.71}   & \textbf{55.82}   \\ \hline
\end{tabular}%
}
\caption{Birds fc6 compression - original accuracy: 55.73\%.}
\label{birdsfc6}
\end{table}

\begin{table}
\resizebox{\linewidth}{!}{%
\begin{tabular}{lcccccc}
\hline
dim kept             & 32     & 64     & 128    & 256    & 512   & 1024  \\
params               & 1.56\% & 3.13\% & 6.25\% & 12.5\% & 25\%  & 50\%  \\ \hline
SVD                  & 26.86 & 44.13 & 52.19 & 54.30 & 54.80 & 55.13 \\ 
SVD + BC             & 28.72 & 47.07 & 53.62 & 54.45 & 55.02 & 55.30 \\ \hline
Pruning (mean)       &  2.49 &  3.45 &  9.15 & 18.31 & 34.05 & 47.24 \\ 
Pruning (max)        &  6.46 &  9.60 & 13.84 & 22.63 & 33.28 & 45.67 \\ \hline
DALR                 & \textbf{51.21}  & \textbf{54.16}  & \textbf{55.21}  & \textbf{55.59}  & \textbf{55.71} & \textbf{55.85} \\ \hline
\end{tabular}%
}
\caption{Birds fc7 compression - original accuracy: 55.73\%.}
\label{birdsfc7}
\end{table}

\begin{table}
\resizebox{\linewidth}{!}{%
\begin{tabular}{lcccccc}
\hline
dim kept             & 32     & 64     & 128    & 256    & 512     & 1024    \\
params               & 0.91\% & 1.82\% & 3.64\% & 7.27\% & 14.54\% & 29.08\% \\ \hline
SVD                  & 14.00 & 47.37 & 59.90 & 75.07 & 77.72 & 78.61 \\
SVD + BC             & 29.55 & 57.93 & 71.96 & 75.91 & 78.00 & 78.63 \\ \hline
Pruning (mean)       &  1.42 &  4.91 & 19.74 & 49.07 & 71.23 & 77.64 \\ 
Pruning (max)        &  1.81 &  4.96 & 10.36 & 33.99 & 60.14 & 74.78 \\ \hline
DALR                 & \textbf{72.17}  & \textbf{76.42}  & \textbf{77.95}  & \textbf{78.22}  & \textbf{78.94} & \textbf{78.94} \\ \hline
\end{tabular}%
}
\caption{Flowers fc6 compression - original accuracy: 78.84\%.}
\label{flowersfc6}
\end{table}

\begin{table}
\resizebox{\linewidth}{!}{%
\begin{tabular}{lcccccc}
\hline
dim kept             & 32     & 64     & 128    & 256    & 512   & 1024  \\
params               & 1.56\% & 3.13\% & 6.25\% & 12.5\% & 25\%  & 50\%  \\ \hline
SVD                  & 58.14 & 70.30 & 75.10 & 77.02 & 78.01 & 78.58 \\
SVD + BC             & 57.07 & 70.14 & 75.53 & 77.15 & 78.00 & 78.50 \\ \hline
Pruning (mean)       & 19.13 & 25.81 & 37.26 & 55.80 & 67.54 & 75.33 \\ 
Pruning (max)        &  8.18 & 18.91 & 27.06 & 42.95 & 62.69 & 72.63 \\ \hline
DALR                 & \textbf{72.32}  & \textbf{76.55}  & \textbf{77.79}  & \textbf{78.48}  & \textbf{78.86} & \textbf{78.86} \\ \hline
\end{tabular}%
}
\caption{Flowers fc7 compression - original accuracy: 78.84\%.}
\label{flowersfc7}
\end{table}

\begin{table}
\resizebox{\linewidth}{!}{%
\begin{tabular}{lcccccc}
\hline
dim kept             & 32     & 64     & 128    & 256    & 512     & 1024    \\
params               & 0.91\% & 1.82\% & 3.64\% & 7.27\% & 14.54\% & 29.08\% \\ \hline
SVD                  & 38.18  & 55.59  & 66.59  & 67.97  & 68.38  & 68.94  \\
SVD + BC             & 46.76  & 60.14  & 66.96  & 68.17  & 68.40  & 69.00  \\ \hline
Pruning (mean)       & 3.98   & 7.27   & 13.70  & 32.14  & 52.02  & 62.46  \\ 
Pruning (max)        & 5.44   & 13.12  & 26.72  & 46.11  & 53.40  & 63.16  \\ \hline
DALR                 & \textbf{64.70}  & \textbf{68.33}  & \textbf{69.31}  & \textbf{69.65}  & \textbf{69.54}   & \textbf{69.52} \\ \hline
\end{tabular}%
}
\caption{Stanford fc6 compression - original accuracy: 68.73\%.}
\label{stanfordfc6}
\end{table}

\begin{table}
\resizebox{\linewidth}{!}{%
\begin{tabular}{lcccccc}
\hline
dim kept             & 32     & 64     & 128    & 256    & 512   & 1024  \\
params               & 1.56\% & 3.13\% & 6.25\% & 12.5\% & 25\%  & 50\%  \\ \hline
SVD                  & 50.98  & 62.27  & 66.72  & 68.08  & 68.44  & 68.60  \\
SVD + BC             & 53.62  & 62.74  & 67.32  & 68.37  & 68.67  & 68.82  \\ \hline
Pruning (mean)       & 12.46  & 17.57  & 25.23  & 34.20  & 51.05  & 61.19  \\ 
Pruning (max)        & 14.84  & 20.37  & 22.90  & 36.35  & 49.30  & 59.33  \\ \hline
DALR                 & \textbf{63.87}  & \textbf{67.46}  & \textbf{68.44}  & \textbf{68.75}  & \textbf{68.78} & \textbf{68.78} \\ \hline
\end{tabular}%
}
\caption{Stanford fc7 compression - original accuracy: 68.73\%.}
\label{stanfordfc7}
\end{table}

\begin{table*}
\resizebox{\textwidth}{!}{%
\begin{tabular}{r|cccccccccccccccccccc|c}
\multicolumn{1}{c}{ } & \rot{\textbf{aeroplane}} & \rot{\textbf{bicycle}} & \rot{\textbf{bird}} & \rot{\textbf{boat}} & \rot{\textbf{bottle}} & \rot{\textbf{bus}} & \rot{\textbf{car}} & \rot{\textbf{cat}} & \rot{\textbf{chair}} & \rot{\textbf{cow}} & \rot{\textbf{diningtable}} & \rot{\textbf{dog}} & \rot{\textbf{horse}} & \rot{\textbf{motorbike}} & \rot{\textbf{person}} & \rot{\textbf{pottedplant}} & \rot{\textbf{sheep}} & \rot{\textbf{sofa}} & \rot{\textbf{train}} & \rot{\textbf{tvmonitor}} & \multicolumn{1}{c}{\textbf{mAP}} \\ \hline
\textbf{No Compression} & 75.9 & 77.4 & 65.3 & 53.9 & 38.0 & 76.8 & 78.2 & 80.9 & 40.6 & 74.0 & 67.2 & 79.4 & 82.4 & 74.9 & 66.2 & 33.4 & 66.0 & 67.3 & 73.3 & 67.1 & 66.9 \\  \hline
\textbf{SVD @ 1024} & 74.4 & 77.6 & 65.9 & 54.9 & 38.4 & 76.7 & 78.2 & 81.6 & 40.0 & 73.0 & 65.9 & 78.9 & 81.9 & 75.7 & 65.6 & 33.7 & 65.3 & 67.3 & 72.4 & 65.7 & 66.6 \\ 
\textbf{SVD + BC @ 1024} & 73.9 & 77.8 & 65.9 & 55.0 & 38.5 & 76.7 & 78.3 & 81.5 & 40.1 & 72.6 & 66.4 & 79.1 & 82.2 & 75.7 & 65.7 & 33.9 & 65.5 & 67.5 & 72.1 & 66.1 & 66.7 \\
\textbf{DALR @ 1024} & 74.6 & 77.7 & 66.4 & 53.7 & 37.6 & 77.0 & 78.1 & 81.7 & 40.5 & 73.3 & 67.4 & 79.7 & 81.9 & 74.8 & 65.9 & 34.0 & 65.9 & 66.9 & 73.7 & 66.3 & 66.9 \\ \hline
\textbf{SVD @ 768} & 74.2 & 77.5 & 65.8 & 53.5 & 37.8 & 76.6 & 78.3 & 82.3 & 39.8 & 72.5 & 66.1 & 79.4 & 81.5 & 74.2 & 65.6 & 33.8 & 64.7 & 67.9 & 71.7 & 66.3 & 66.5 \\ 
\textbf{SVD + BC @ 768} & 74.4 & 77.5 & 65.6 & 53.8 & 38.0 & 77.3 & 78.2 & 82.2 & 39.7 & 72.5 & 66.3 & 79.7 & 82.1 & 75.4 & 65.7 & 34.0 & 65.1 & 68.1 & 71.7 & 66.0 & 66.7 \\ 
\textbf{DALR @ 768} & 74.2 & 77.5 & 66.7 & 53.8 & 37.8 & 77.4 & 78.1 & 82.0 & 40.1 & 72.7 & 66.2 & 79.0 & 81.7 & 75.5 & 66.1 & 33.9 & 65.8 & 66.7 & 73.2 & 66.0 & 66.7 \\ \hline
\end{tabular}%
}
\caption{Compression and bias compensation results on Fast-RCNN on PASCAL 2007.}\label{tab_pascal}
\end{table*}

In the previous experiment we evaluated the layer compression separately for fc6 and fc7. To get a better understanding of the potential joint compression, we perform a compression experiment with DALR on both layers simultaneously. In order to find suitable compression pairs for both layers at the same time, we implemented an iterative solution. At each step, we slightly increase the compression on both layers. Then, both options are evaluated on the validation set, and the compression rate with better performance is applied to the network and used on the next iteration. When both steps in compression exceed a defined drop in performance (here set to a 1\% accuracy drop), the iterative process stops and the compressed network is evaluated on the test set. Results are shown in Table~\ref{iterativePip}. This implementation tends to compress fc6 more because fc6 has more room for compression than fc7, as seen also in the experiments reported in Tables~\ref{birdsfc7} to~\ref{stanfordfc7}. The results show that we can compress the parameters of the fully connected layers to as few as 14.88\% for Flowers, as few as 6.81\% for Birds, and as few as 29.85\% for Stanford while maintaining close to optimal performance. 

\begin{table}
\resizebox{\linewidth}{!}{%
\begin{tabular}{lccccc}
\hline    & Orig. Acc & Compr. Acc & fc6 red. & fc7 red. & total red. \\ \hline
Birds     & 55.73 & 55.97 &  4.66\% & 19.97\% &  6.81\% \\
Flowers   & 78.84 & 77.62 & 10.45\% & 41.99\% & 14.88\% \\
Stanford  & 68.73 & 69.43 & 25.59\% & 55.96\% & 29.85\% \\ \hline
\end{tabular}%
}
\caption{Reduction in number of parameters for both fc6 and fc7.}
\label{iterativePip}
\end{table}

\subsection{Object detection}
One of the most successful approaches to object detection is RCNN~\cite{RCNN} (including its Fast~\cite{girshick2015fast} and Faster~\cite{ren2015faster} variants). This approach is also an example of the
effectiveness of the fine-tuning approach to domain transfer, and also of the importance of network compression for efficient detection. The authors of \cite{girshick2015fast} analyzed the timings of forward layer computation and found that 45\% of all computation time was spent in fc6 and fc7. They then applied truncated SVD to compress these layers to 25\% of their original size.  This compression however came with a small drop in performance of 0.3 mAP in detection on PASCAL 2007.

For comparison, we have also run SVD with bias compensation and our compression approach based on low-rank matrix decomposition. Results are presented in Table~\ref{tab_pascal}. Here we apply compression at varying rates to fc6 (which contains significantly more parameters), and compress fc7 to 256 pairs of basis vectors (which is the same number used in~\cite{girshick2015fast}).  What we see here is that at the same compression rate for fc6 (1024) proposed in~\cite{ren2015faster}, our low-rank compression approach does not impact performance and performs equal to the uncompressed network. When we increase the compression rate of fc6 (768) we see a drop of 0.4\% mAP for standard SVD and only half of that for both SVD with bias compensation and DALR.

\section{Conclusions and discussion}
\label{sec:conclusions}
We proposed a compression method for domain-adapted networks. Networks which are designed to be optimal on a large source domain are often overdimensioned for the target domain. We demonstrated that networks trained on a specific domain tend to have neurons with relatively flat activations rates, indicating that almost all neurons are equally important in the network. However, after transferring to a target domain, activation rates tend to be skewed. This motivated us to consider activation statistics in the compression process. We show that compression which takes activations into account can be formulated as a rank-constrained regression problem which has a closed-form solution. As an additional contribution we show how to compensate the bias for the matrix approximation done by SVD. This is consistently shown to obtain improved results over standard SVD.  

Experiments show that DALR not only removes redundancy in the weights, but also balances better the parameter budget by keeping useful domain-specific parameters while removing unnecessary source-domain ones, thus achieving higher accuracy with fewer parameters, in contrast to truncated SVD, which is blind to the target domain. On further experiments in image recognition and object detection, the DALR method significantly outperforms existing low-rank compression techniques. With our approach, the fc6 layer of VGG19 can be compressed 4x more than using truncated SVD alone -- with only minor or no loss in accuracy. 

The Bias Compensation and DALR techniques were applied to fully connected layers in this work. To show the effectiveness of those methods we applied them to standard networks with large fully connected layers. On more recent networks, like ResNets~\cite{he2016deep}, most of the computation has moved to convolutional layers, and the impact of the proposed method would be restricted to the last layer. However, VGG-like networks are very much used in current architectures~\cite{bansal2017pixelnet,zhou2016places,redmon2016you}. Extending the proposed compression method to convolutional layers is an important research question which we aim to address in future works.

Our paper shows that domain transferred networks can be significantly compressed. The amount of compression seems to correlate with the similarity~\cite{azizpour2016factors,yosinski2014transferable} between the source and target domain when we compare it to the ordering proposed in~\cite{azizpour2016factors} (see Table II). According to this ordering, the similarity with respect to ImageNet in descending order is image classification (PASCAL), fine-grained recognition (Birds, Flowers) and compositional (Stanford). We found found that higher compression rates can be applied in target domains further away from the source domain.

\minisection{Acknowledgements}
We thank Adri\`{a} Ruiz for his advice on optimization. Herranz acknowledges the European Union's H2020 research under Marie Sklodowska-Curie grant No. 6655919. Masana acknowledges 2017FI-B-00218 grant of Generalitat de Catalunya, and their CERCA Programme. We acknowledge the  Spanish project TIN2016-79717-R, the CHISTERA project M2CR (PCIN-2015-251). We also acknowledge the generous GPU support from Nvidia.
{\small
\bibliography{ICCV2017_CameraReady}
\bibliographystyle{ieee}
}

\end{document}